\documentclass[conference]{IEEEtran}
\IEEEoverridecommandlockouts
\usepackage{cite}
\usepackage{amsmath,amssymb,amsfonts}
\usepackage{algorithmic}
\usepackage{graphicx}
\usepackage{textcomp}
\usepackage{xcolor}
\def\BibTeX{{\rm B\kern-.05em{\sc i\kern-.025em b}\kern-.08em
    T\kern-.1667em\lower.7ex\hbox{E}\kern-.125emX}}
\usepackage{graphicx}
\usepackage{comment}
\usepackage{todonotes}
\usepackage[algo2e,linesnumbered,ruled,vlined]{algorithm2e}
\begin{document}

\title{FLICU: A Federated Learning Workflow for Intensive Care Unit Mortality Prediction\\
%{\footnotesize \textsuperscript{*}Note: Sub-titles are not captured in Xplore and should not be used}%\thanks{Identify applicable funding agency here. If none, delete this.}
}

\author{\IEEEauthorblockN{1\textsuperscript{st} Lena Mondrejevski}
\IEEEauthorblockA{\textit{Dept. of Computer  \& Systems Sciences,} \\
\textit{Stockholm University;} \\
%\textit{Research \& Business Development, Getinge}\\
%\textit{QRIOS IT}\\
\textit{Getinge; QRIOS IT}\\
Stockholm, Sweden \\
lena.mondrejevski@dsv.su.se}
\and
\IEEEauthorblockN{2\textsuperscript{nd} Ioanna Miliou}
\IEEEauthorblockA{\textit{Dept. of Computer  \& Systems Sciences,} \\
\textit{Stockholm University}\\
Stockholm, Sweden \\
ioanna.miliou@dsv.su.se}
\and
\IEEEauthorblockN{3\textsuperscript{rd} Annaclaudia Montanino}
\IEEEauthorblockA{\textit{Research \& Business Development,} \\
\textit{Getinge}\\
Stockholm, Sweden \\
annaclaudia.montanino@getinge.com}
\and
\IEEEauthorblockN{4\textsuperscript{th} David Pitts}
\IEEEauthorblockA{\textit{Research \& Business Development,} \\
\textit{Getinge}\\
Stockholm, Sweden \\
david.pitts@getinge.com}
\and
\IEEEauthorblockN{5\textsuperscript{th} Jaakko Hollmén}
\IEEEauthorblockA{\textit{Dept. of Computer  \& Systems Sciences,} \\
\textit{Stockholm University}\\
Stockholm, Sweden \\
jaakko.hollmen@dsv.su.se}
\and
\IEEEauthorblockN{6\textsuperscript{th} Panagiotis Papapetrou}
\IEEEauthorblockA{\textit{Dept. of Computer  \& Systems Sciences,} \\
\textit{Stockholm University}\\
Stockholm, Sweden \\
panagiotis@dsv.su.se}
}

\maketitle              

\begin{abstract}
Although Machine Learning (ML) can be seen as a promising tool to improve clinical decision-making for supporting the improvement of medication plans, clinical procedures, diagnoses, or medication prescriptions, it remains limited by access to healthcare data. Healthcare data is sensitive, requiring strict privacy practices, and typically stored in data silos, making traditional machine learning challenging. Federated learning can counteract those limitations by training machine learning models over data silos while keeping the sensitive data localized. This study proposes a federated learning workflow for ICU mortality prediction. Hereby, the applicability of federated learning as an alternative to centralized machine learning and local machine learning is investigated by introducing federated learning to the binary classification problem of predicting ICU mortality. We extract multivariate time series data from the MIMIC-III database (lab values and vital signs), and benchmark the predictive performance of four deep sequential classifiers (FRNN, LSTM, GRU, and 1DCNN) varying the patient history window lengths (8h, 16h, 24h, 48h) and the number of FL clients (2, 4, 8). The experiments demonstrate that both centralized machine learning and federated learning are comparable in terms of AUPRC and F1-score. Furthermore, the federated approach shows superior performance over local machine learning. Thus, the federated approach can be seen as a valid and privacy-preserving alternative to centralized machine learning for classifying ICU mortality when sharing sensitive patient data between hospitals is not possible. %Directions for future work include interpretability methods to understand the most influential features in predicting the risk of ICU mortality in a federated learning setup.
\end{abstract}


\section{Introduction}
\label{sec:introduction}
Healthcare generates a vast amount of data that, if adequately leveraged, has the potential to lead to improved clinical decision-making even at the single patient level. This potential is, however, yet to be fully realized. Machine Learning (ML) is a promising tool to make a step towards this goal, as it can achieve higher predictive performance against current conventional approaches for several clinical prediction tasks
%, for example, medication plans, clinical procedures, diagnosis, and medication prescription
\cite{purushotham2018benchmarking,harutyunyan2019multitask,shamout2019deep}. 
However, when it comes to accessing healthcare data, traditional ML faces several limitations.
%Although ML is promising, it remains limited by access to healthcare data.. 
Due to their sensitive nature, patient data is usually stored in data silos and protected by legal and ethical practices. As a result, ML models could be trained on individual small local datasets. Nevertheless, this Local Machine Learning (LML) approach makes it challenging to obtain models that are generalizable enough, %to make predictions on single patient cases
as those local datasets are generally biased and/or too limited.
The standard approach to secure access to more extensive datasets is to anonymize, extract, and aggregate data from multiple healthcare institutions and train ML models centrally, outside the hospital premises. The advantage of this type of Centralized Machine Learning (CML) over LML is that the obtained models are more generalizable, as they are based on data from several healthcare institutions. % and thus be more robust. 
However, this approach comes with heavy restrictions and several limitations in terms of scalability, security, cost-efficiency, and data privacy. For example, even anonymized data, when shared, can impose risks to patient privacy \cite{sweeney2000simple}. Thus, data is often required to remain inside the hospital premises. 

\begin{figure*}[t]
\centering
\includegraphics[width=0.9\textwidth]{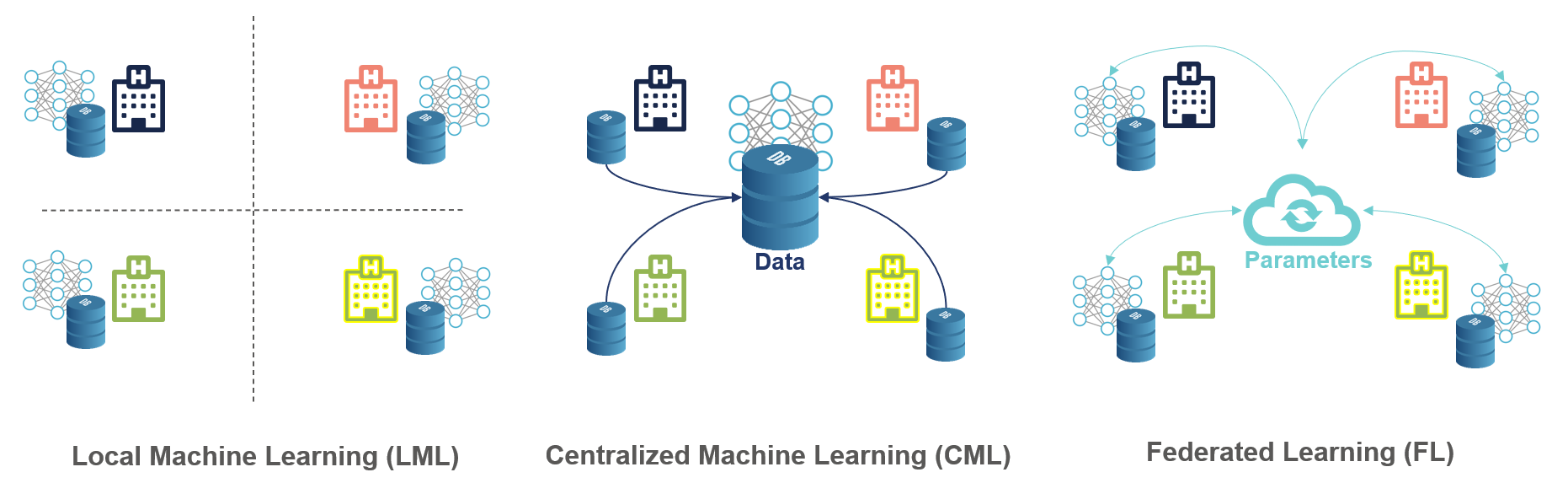}

\caption{Comparison of LML, CML, and FL.} \label{fig:fig_fl_lml_cml}
\vspace{-2mm}
\end{figure*}

Counteracting the previously mentioned limitations of CML and LML, Federated Learning (FL) \cite{mcmahan2017communication,kairouz2019advances} trains the models over data silos while keeping the sensitive data localized (see Fig. \ref{fig:fig_fl_lml_cml}). Its distributed design ensures that data is not shared between clients, for example, hospitals, but instead, only the local model parameters are shared, which are subsequently aggregated to a joint model. Thus, FL can be seen as a propitious solution for privacy-preserving ML within healthcare. An additional advantage of FL is its capability to be seamlessly integrated with existing electronic healthcare systems storing valuable data, like Electronic Health Records (EHRs) \cite{xu2021federated,pfitzner2021federated}.

%It has been shown that traditional ML can leverage the vast amount of healthcare data and improve the current conventional approaches with higher predictive performance for several clinical prediction tasks \cite{purushotham2018benchmarking,harutyunyan2019multitask,shamout2019deep}. 
One of the most researched clinical prediction tasks where ML has been applied, is predicting the probability of patient death during hospitalization \cite{fang2020early}.
%Scoring systems are currently used for this scope, which are able to classify and stratify patients by their severity of illness \cite{brandenburg2014hospital}.
The unit where this need is more prominent is arguably the Intensive care unit (ICU), since this is the unit where the patients with the most severe and life-threatening medical conditions %sickest and most complex patients 
are admitted and cared for. As a result, the ICU is often the unit with the highest mortality rate. \emph{ICU mortality} is defined as death during an ICU stay \cite{brandenburg2014hospital}. While the conventional way of mortality risk assessment is scoring systems, which are able to classify and stratify patients by their severity of illness \cite{brandenburg2014hospital}, several traditional ML-based solutions have been recently proposed. For example, Johnson and Mark \cite{johnson2017real} focus on real-time ICU mortality prediction using logistic regression and gradient boosting, while Pattalung and Chaichulee \cite{pattalung2021comparison} compare multiple ML algorithms for ICU mortality prediction. Additionally, Pattalung et al. \cite{pattalung2021feature} focus on predicting the risk of ICU mortality by combining Recurrent Neural Networks (RNNs) with interpretable explanations. Finally, Rinta-Koski et al. \cite{rinta-koski-2018} propose Gaussian process classification for mortality prediction in a neonatal ICU. 
%in-hospital mortality in the context of a neonatal ICU.
%Predictive scoring systems have been historically developed to objectively measure severity of disease and the prognosis of the patient in the intensive care unit (ICU). Such information is helpful not only for clinical decision-making, but also for standardizing research, and comparing the quality of patient care across the ICUs. 
%The unit with the highest mortality rate in any hospital is the Intensive Care Unit (ICU), due to the severity and complexity of the admitted patient cases \cite{brandenburg2014hospital}. One way to determine the severity of a disease and come up with a prognosis of the patient outcome is by using predictive scoring systems \cite{brandenburg2014hospital}.
%Ideally, scoring systems are well-calibrated, have high-level discrimination, are applicable to all patient populations, and can be used in different countries \cite{rapsang2014scoring}. However, due to the large variability in ICU patient populations, traditional scoring systems, such as Acute Physiology Age Chronic Health Evaluation (APACHE), % or Simplified Acute Physiology Score (SAPS) %, Sequential Organ Failure Assessment (SOFA)-
%cannot be used to make predictions on individual cases. Moreover, such scoring systems typically assess the likelihood of in-hospital mortality, while ICU mortality is arguably of higher interest for ICU practitioners \cite{bouch2008severity}.
%The ICU is a highly specialized medical environment where critically ill patients are monitored and assisted around the clock.

Moreover, there are several recent FL solutions for \emph{in-hospital mortality} prediction \cite{lee2020federated,budrionis2021benchmarking}, which is defined as patient death during a hospital stay after being admitted to an ICU. Lee and Shin \cite{lee2020federated} demonstrated that FL can reach a comparable predictive performance to that of CML in predicting in-hospital mortality using a standard Long Short-Term Memory (LSTM). The authors compare the performance of CML and FL in a simple experimental FL setup with three clients and observe the influence on the performance of balanced and imbalanced distribution of data (amount not labels) amongst clients. Budrionis et al. \cite{budrionis2021benchmarking} extended the work of Purushotham et al. \cite{purushotham2018benchmarking}, who benchmark deep learning algorithms to more traditional ML algorithms on MIMIC-III. They compare the performance of CML and FL more extensively than Lee et al. \cite{lee2020federated} with experiments in a more realistic deployment setting of FL, studying the influence of the number of clients, amount of data, and data distribution on predictive performance and inference and training duration.

Despite the FL-based solutions for in-hospital mortality prediction mentioned above, little emphasis has been given to ICU mortality prediction using FL. This paper addresses this limitation by:
%and employ several deep learning models in an FL setup on MIMIC-III. 
%inspired by the work of Pattalung et al. \cite{pattalung2021feature}, we employ several ML models in an FL setup on MIMIC-III for predicting ICU mortality. 
%The models are benchmarked in terms of predictive performance (AUPRC and F1-score) 
%trained in a centralized vs. a local vs. a federated setting.
(1) proposing FLICU, a workflow for retrospective analysis of ICU mortality using FL alongside sequential deep neural network classifiers;
    %Federated learning is applied to the problem of predicting ICU mortality.
(2) comparing the proposed FL solution against LML and CML in terms of predictive performance using an extract from the MIMIC-III database;
(3) benchmarking four common sequential neural network architectures (1DCNN, FRNN, LSTM, and GRU) as parts of our workflow for different patient history window lengths (8h, 16h, 24h, and 48h before the discharge/death event in the ICU);
    %\item We further investigate the effect of using different window widths of the considered temporal data (8h, 16h, 24h, and 48h) of ICU patients before the death event in the ICU.
    %the role of multivariate time series data on the predictive performance of the federated learning models. This includes the effect of using different window widths of the considered temporal data (8h, 16h, 24h, and 48h) of ICU patients before the death event in the ICU.
(4) studying the sensitivity of the four FL models as the number of FL clients varies (2, 4, and 8).

% Organization of the paper, section by section:
%The rest of the paper is organized as follows: after the introductory remarks in Section \ref{sec:introduction}, we define the problem in Section \ref{sec:formulation} and our proposition in Section \ref{sec:flicu}. The Section \ref{sec:empirical_eval} describes the empirical evaluation and the results. In Section \ref{sec:summary_conclusions}, we summarize our findings and conclude the paper.

\section{Problem Formulation}
\label{sec:formulation}
The problem studied in this paper can be formulated as a binary classification problem for ICU mortality prediction, where the label indicates whether a patient died during an ICU stay or got discharged.
%A global view of ICU mortality is considered, which includes all patients and not only patients with a specified disease. 
Given a set of $p$ ICU patients, we define $\mathcal{D}$ to be a collection of multivariate time series variables, with $|\mathcal{D}|=p$. Each $\mathcal{D}_i\in \mathcal{D}$ describes a set of vital signs and lab tests of the $i^{th}$ ICU patient over time and contains $q$ univariate time series, with $|\mathcal{D}_i|=q$. The dimensionality (i.e., length) of the $j^{th}$ time series $\mathcal{D}_{ij} \in \mathcal{D}_i$ may vary due to the variable sampling rates used for collecting the vital signs and lab values. 

The objective of the CML approach is to learn a function $h_{CML}(\cdot)$ using $\mathcal{D}$, such that, given an ICU patient, it assesses whether the patient will die during an ongoing ICU admission. Moreover, consider a set of $K$ clients, each having its own patient cohort $\mathcal{D}^k\subset \mathcal{D}$, such that $\mathcal{D}^a\cap\mathcal{D}^b = \emptyset, \forall (a, b) \in [1, K]\times[1, K]$. For LML, the goal is to build a set of classifiers $h_{LML}^k(\cdot)$, using $\mathcal{D}^k,  \forall k \in[1,K]$, for solving the binary classification task of ICU mortality prediction locally at each client. 

For FL, we build a local classifier $h_{FL}^k(\cdot)$ using $\mathcal{D}^k$ for each client $k \in[1,K]$ and denote with $w^k$ the set of the local weights learned by $h_{FL}^k(\cdot)$. Our objective is to define a global classifier $h_{FL}(\cdot)$ that is learned as a function of the local weights, i.e., $h_{FL}(\{w^1, \ldots, w^K\})$, that optimizes its weight configuration without sharing the local datasets. For example, $h_{FL}(\cdot)$ can be a weighted average of the local weights.

\section{FLICU: A Federated Learning Workflow for Intensive Care Unit Mortality Prediction}
\label{sec:flicu}

The proposed workflow comprises three steps: (1) feature extraction within a time window, (2) local FL model training, and (3) global FL model training.

\subsection{Feature Extraction within a Time Window}
For each ICU patient $i$, we identify one of two critical time points, i.e., the time of death in the ICU (positive class) or the time of discharge from the ICU (negative class). Given a fixed time window $W$ (8h, 16h, 24h, and 48h), for each $\mathcal{D}_{ij} \in \mathcal{D}_i$, and based on Pattalung et al. \cite{pattalung2021feature}, we only consider the medical events that occurred within $W$ hours before the last recorded vital sign or lab value. 
Knowing the time of ICU death/discharge, we assume that the most important information about the critical event is at the end of the ICU stay, and we want to explore how much of this information (window size) is really relevant based on the predictive performance achieved by the model.

For each $W$, we extract vital signs and lab values. As vital signs and lab values have different temporal characteristics (0.5–1.5 vital signs per hour, 1-2 lab values per 8 hours), the variables corresponding to vital signs are re-sampled in 1h time intervals, with mean as aggregation function, while lab value variables are re-sampled in 8h time intervals. Missing values are imputed using linear interpolation, as eliminating patients can bias the study. Furthermore, if certain variables are never observed for a patient, their values are set to the variable mean.

\begin{comment}
\begin{figure}
\centering
\includegraphics[width=0.49\textwidth]{figures/Windowing approach.PNG}
\caption{Windowing Approach.} \label{fig_windowing_approach}
\end{figure}
\end{comment}

\subsection{Local FL Model Training}
%Let $\mathcal{D}^k$ be a local training set of a patient cohort in client $k$. The objective is to find for each model the optimal network weight configuration $\Theta_k$ that minimizes the prediction error on $\mathcal{D}_k$,  i.e., $f_k(w) = \min\limits_{w \in \mathbb{R}^{d}} L(\mathcal{D}_k,w)$, where $L(\cdot)$ is the loss function of the network and $w$ is the set of network weights.

%As the local model learning is based on traditional deep learning, the learning objective is $\min\limits_{w \in \mathbb{R}^{d}}f(w)$, where $f(w)=\frac{1}{n}\sum_{i=1}^{n}l(x_{i}, y_{i}, w)$ for a training dataset that contains $n$ samples $(x_{i}, y_{i})$, $i \in [1,n]$, and loss $l(x_{i}, y_{i}, w)$ of the prediction of example $(x_{i}, y_{i})$ with model parameters $w$ \cite{mcmahan2017communication}.

In this study, we explore four neural network architectures. We first use a one-dimensional convolutional neural network (1DCNN),  which creates a convolution kernel that is convolved with the input layer over one dimension. Additionally, we use three sequential deep learning architectures: a Fully-connected RNN (FRNN) architecture \cite{bengio1994learning}, %RNNs are neural networks with 'internal memory' as connections between nodes form a sequence. Hence, they consider the current input but also what has been learned from previous inputs, making RNNs very well suited for sequential data like time series. Nevertheless, RNNs have the deficit of short-term memory due to difficulties in learning long-term dependencies \cite{bengio1994learning}. Thus, additional RNN architectures are investigated.
and its two adaptations, i.e., a Long Short-Term Memory (LSTM) architecture \cite{hochreiter1997long}, and a Gated Recurrent Unit (GRU) architecture \cite{cho2014properties}. %The key difference between both RNN adaptions is their internals, as LSTM has three types of gates, i.e., input, output, and forget gates, while GRU has two gates, i.e., reset and update gates. Usually, LSTM is preferred for longer data sequences.
%The main difference between CNNs and RNNs is their ability to process sequential (for example, temporal) information. RNNs are designed for the fact, whilst CNNs are designed to exploit spatial information and are less capable of effectively interpreting temporal information.

For each client $k \in[1,K]$, the local FL models $h_{FL}^k(\cdot)$ are trained using their local cohort $\mathcal{D}^k$, and each set of local weights $w^k$ is then passed to the central server. The local model consists of two parallel input channels (one for vitals and one for labs), with one Conv1D layer followed by one Flatten layer (kernel sizes 1 and 8) each, for the 1DCNN model, or with 3 RNN layers of 16 units each for the three sequential neural network classifiers (according to \cite{pattalung2021feature}). For all classifiers, we perform batch normalization sub-sequentially, followed by a concatenation of the outputs and fusion of the concatenated outputs via two fully connected layers. We calculate the final outputs ([0,1]) with a Sigmoid layer computing the risk of death in the ICU or ICU discharge. Moreover, we use Adaptive Moment Estimation (Adam) optimization \cite{kingma2014adam} and binary cross-entropy loss function, which is suitable for binary classification tasks.

%On the one hand, RNNs employ `internal memory' by considering both the current input and what has been learned from previous inputs; hence they are suitable for sequential data. Nevertheless, they have the deficit of short-term memory due to their difficulty in learning long-term dependencies. On the other hand, by learning what information is relevant to make predictions over time, both LSTM and GRU are capable of learning long-term dependencies \cite{bengio1994learning}.

\subsection{Global FL Model Training}
We consider $K$ clients, each having its own local patient cohort $\mathcal{D}^k, k\in[1,K]$. Initially, centrally (i.e., in the central server), we initialize a global model that is shared with all the clients, whereby each client represents one hospital. Subsequently, we conduct several FL rounds: all clients train their local models on their local data based on the received global model (in defined epochs and mini-batch size), then, the clients send the resulting local model to the global server, and the server aggregates the local models and updates the global model with the aggregated results. The FL rounds are repeated until a defined stopping criterion is met. 

\begin{comment}
\begin{figure*}
\centering
\includegraphics[width=0.9\textwidth]{figures/FL Setup.PNG}
\caption{Federated Learning Setup.} \label{fig_fl_setup}
\end{figure*}
\end{comment}

The objective for FL is to optimize the global model parameters using an aggregation of the local model parameters. This is accomplished by minimizing the following function $f_{FL}(w) = \sum_{k=1}^{K}\frac{n_{k}}{n}f_{FL}^{k}(w)$,
with $f_{FL}^{k}(w)=\frac{1}{n_{k}}L(\mathcal{D}^k, w)$, $n_k=|\mathcal{D}^k|$, and $w$ denoting the global model weights \cite{mcmahan2017communication}.

Our FL model training approach is based on the Federated Averaging (FedAvg) algorithm \cite{mcmahan2017communication}. FedAvg is most commonly used for FL with neural networks, an algorithm based on iteratively averaging the stochastic gradient descent (SDG) weights generated locally. It has been shown that FedAvg is robust for non-IID and imbalanced data distributions \cite{mcmahan2017communication}, which is very common for medical data. In FedAvg, the local models are updated multiple times (multiple batch gradient calculations) before sending the model weights back to the server for aggregation, contrary to Federated SGD (FedSGD), where a single step of gradient descent is performed per client in each FL round.

\begin{algorithm2e}%[H]
\SetAlgoLined
\KwResult{Optimized global model } \label{alg_fed_avg}
 initialize $w_0$\;
 \While{stopping criterion not met}{
    \ForEach{client $k\in[1,K]$}{
        download $w$ from central server\;
        $w^k\gets w$\;
        \ForEach{mini-batch}{
            $w^k\gets w^k-\eta g^{k}$\;
        }
        return $w^k$ to central server\;
    }
    update $w\gets \sum_{k=1}^{K}\frac{n_{k}}{n}w^k$ in central server\;
 }
 \caption{Global FL Model Training}
\end{algorithm2e}

The pseudocode of our approach is presented in Algorithm \ref{alg_fed_avg}, inspired by the FedAvg algorithm.
%The FedAvg algorithm can be described more formally \cite{mcmahan2017communication}: 
In line 1, we initialize the global model weights $w_0$ in the central server.
Then, the central server shares the current global model weights with the selected clients at the beginning of each FL round (lines 4 and 5). Subsequently, each client $k$ computes the gradient $g^{k}=\nabla f_{FL}^{k}(w^k)$ using its local data and performs local updates with fixed learning rate $\eta$ (line 7) in multiple iterations (dependent on mini-batch size), resulting in $w^k$. Finally, the weights are returned to the central server (line 9), and the results are aggregated and updates applied to the central server (line 11). The previously described steps are iterated for several FL rounds.

\begin{comment}
\begin{equation}
    w_{t+1}\gets \sum_{k=1}^{K}\frac{n_{k}}{n}w_{t+1}^k
\end{equation}
\end{comment}

In our FedAvg setup, all clients perform computations on each FL round, as this is more suitable for the domain. Each client performs 1 training pass over the local dataset per FL round, and the local mini-batch size for client updates is dependent on the number of clients $K$ ($64/K$).

\section{Empirical Evaluation}
\label{sec:empirical_eval}

\subsection{Dataset}
We use the MIMIC-III dataset \cite{johnson2016mimic} for this study, which is a publicly available critical care database containing de-identified clinical data of patients admitted to an ICU at the Beth Israel Deaconess Medical Center (BIDMC) from 2001 to 2012.
We follow the approach described in Pattalung et al. \cite{pattalung2021feature} for the pre-processing and feature extraction steps, expanding on the publicly available code of the mimic-code GitHub repository \cite{mimic-code,johnson2018mimic}. We extract patient demographic information for pre-processing and labeling (icustay id, first icu stay, first careunit, length of stay icu, deathtime icu), as well as statistical purposes (gender, ethnicity, admission age), which are presented briefly in Tables \ref{table_demographics_1} and \ref{table_demographics_2}. Note that patients older than 89 have age values of 300 in MIMIC-III due to privacy reasons, which are set to 90 in this study to reflect reality more closely. Additionally, we extract patients' vital signs and lab values, collected during their ICU stays for modeling. We obtain 23 ICU mortality clinically relevant variables, which are in the form of time series \cite{pattalung2021feature}, 7 vital signs (heartrate, systolic blood pressure, diastolic blood pressure, mean blood pressure, respiratory rate, temperature, peripheral oxygen saturation), and 16 lab values (albumin, blood urea nitrogen, bilirubin, lactate, bicarbonate, band neutrophi, chloride, creatinine, glucose, hemoglobin, hematocrit, platelet count (platelet), potassium, partial thromboplastin time, sodium, white blood cells). Additionally, we prune data outliers and perform grouping of similar clinical variables (using the item ids, see \cite{mimic-code}). Finally, we filter the patients, following the steps described below, resulting in a total of 19414 patients:
\begin{enumerate}
    \item Filter for the first ICU stay of each patient.
    \item Exclude patients admitted to the Neonatal Intensive Care Unit (NICU) and Pediatric Intensive Care Unit (PICU).
    \item Filter for patients whose length of stay in the ICU was at least 48h to ensure sufficient data for analysis.
    \item Filter for patients for which observations (vital signs and laboratory values) are recorded for at least 48h.
\end{enumerate}

%\todo{Add vitals and lab statistics to the appendix}
\begin{table}
\centering
\caption{Patient demographics of final study cohort.}\label{table_demographics_1}
\vspace{-2mm}
\begin{tabular}{|l|l|l|l|l|}
\hline
Demographics & Total & Death & Survival\\
\hline
\hline
Patients & 19414 & 1892 & 17522\\
{\bfseries Gender} & & &\\
Female & 8582 & 879 & 7703\\
Male & 10832 & 1013 & 9819\\
{\bfseries Ethnicity} & & &\\
Caucasian & 13706 & 6021 & 7685\\
African American & 1447 & 801 & 646\\
Asian & 445 & 174 & 271\\
Hispanic/Latino & 604 & 249 & 355\\
Others/Unknown & 3212 & 1337 & 1875\\
\hline
\end{tabular}
\vspace{-2mm}
\end{table}

\begin{table}
\centering
\caption{Admission age and length of stay of final study cohort.}\label{table_demographics_2}
\vspace{-2mm}
\begin{tabular}{|l|ccc|ccc|}
\hline
 & \multicolumn{3}{|c|}{\textbf{Admission age (years)}} & \multicolumn{3}{|c|}{\textbf{Length of 1st ICU stay (days)}}\\
\hline
 & Total & Death & Survival & Total & Death & Survival\\
\hline
\hline
Count & 19414 & 1892 & 17522 & 19414 & 1892 & 17522 \\
Mean & 64.83 & 68.56 & 64.42 & 6.82 & 9.46 & 6.53 \\
Std & 17.09 & 16.15 & 17.14 & 7.50 & 8.90 & 7.28 \\
Min & 15.19 & 16.47 & 15.19 & 2.00 & 2.01 & 2.00 \\
Max & 90.00 & 90.00 & 90.00 & 153.93 & 97.30 & 153.93 \\
\hline
\end{tabular}
\vspace{-2mm}
\end{table}

%"To keep the personal information from leakaging and ensure similar experimental settings with related works, we only use the first ICU admissions of each patient." - This is a reason for using only the first ICU admission. However, this is more valid when there is demographic data included, I think!

Labels are assigned to each unique patient. Patients that died during the ICU stay are included in the positive group (label = 1). Patients being alive throughout the entire ICU stay, up until ICU discharge, are included in the negative group (label = 0). The labeling process resulted in 1892 patients (9.75\%) in the positive class and 17522 (90.25\%) in the negative class, which demonstrates the high class imbalance of the dataset.

\begin{comment}
\begin{table}
\centering
\caption{Data distribution of labels.}\label{table_class_distribution}
\begin{tabular}{|l|l|l|}
\hline
Class & Patient count & Patient fraction \\
\hline
Death during ICU stay (Positive) & 1889 & 9.75\% \\
Survived ICU stay (Negative) & 17486 & 90.25\% \\
Total & 19375 & 100\% \\
\hline
\end{tabular}
\end{table}
\end{comment}

\subsection{Evaluation Strategy}
We evaluate LML, CML, and FL on the same testing data splits using 5-fold cross-validation to eliminate randomness induced by dataset partitioning. %(see Figure \ref{fig:fig_eval_framework}). 
All folds consist of 20\% of the whole data each, whereby each fold serves as a testing set once. The remaining four folds are again split into 85\% training and 15\% validation set (CML approach). In the FL approach, the remaining data is firstly split into $K$ cohorts (one per client), then each client's cohort is also split into 85\% training and 15\% validation set. We use a stratified sampling process for each splitting procedure because the data is highly imbalanced. To further ensure comparability, each neural network type is initialized with the same random weights to ensure the same starting point for the optimization in all approaches.

\begin{comment}
\begin{figure*}[t]
\centering
\includegraphics[width=0.9\textwidth]{figures/Evaluation Framework.PNG}
\caption{Evaluation framework - Data splitting of CML and FL ($K$=[2,4,8]) into training, validation and testing datasets within each 5-fold Cross Validation iteration.} \label{fig:fig_eval_framework}
\end{figure*}
\end{comment}

In all three approaches, CML, LML, and FL, firstly, we normalize the training and validation datasets. The class imbalance is taken into consideration by using class weights during model training by giving both positive and negative classes equal importance on gradient updates. 
%Additionally, the output bias is initialized to reflect the class imbalance and help initial convergence ($b_0 = log_e(pos/neg)$). 
We train the CML models on a mini-batch size of 64 and the LML models (2, 4, and 8 clients) on a mini-batch size of $64/K$ using a maximum of 100 epochs and an initial learning rate of 0.01, which decreases by 50\% every 5 epochs to avoid undesirable divergent behavior in the loss function. Furthermore, we use early stopping via monitoring the validation loss with a patience value of 30. When reaching this criterion, we restore the weights of the epoch with the best validation results and test the final model on the normalized testing dataset.

We train all FL clients' local models in 1 epoch to maintain a high training speed. This is considerably lower than the CML, and LML approaches due to the iterative averaging process, a mini-batch size of $64/K$, and a maximum of 100 FL rounds (similar to max epochs in CML). As in CML, the initial learning rate is set to 0.01 for all clients. Each training round follows the aggregation of weights with FedAvg. We pass the global model with averaged weights to the clients and evaluate them locally on their validation dataset. We monitor the average validation loss as FL stopping criterion, with the patience set to the same number as in CML and LML: if the client's averaged validation loss does not improve over 30 rounds, we initiate early stopping. Eventually, we restore the FL model with the lowest loss and test it on the normalized test dataset. Finally, we repeat this process for 2, 4, and 8 clients.

\subsection{Results and Discussion}
We tackle the task of predicting ICU mortality using multiple sequential classifiers in a federated setting on the MIMIC-III dataset. We evaluate the predictive performance of 1DCNN and three types of RNNs (FRNN, LSTM, and GRU), varying the patient history window lengths (8h, 16h, 24h, and 48h) and the number of FL clients (2, 4, and 8). Additionally, we compare the results of the three approaches, FL, LML, and CML. In Table \ref{tab_results}, we comparatively present the performance of all combinations using the following evaluation metrics: AUPRC and F1-Score.

\begin{table*}[t]
\scriptsize
\centering
\caption{Predictive Performance of 1DCNN, FRNN, LSTM, and GRU.}\label{tab_results}
\vspace{-2mm}
\begin{tabular}{|l|cccc|cccc|}
\hline
 & \multicolumn{4}{|c|}{\textbf{AUPRC}} & \multicolumn{4}{|c|}{\textbf{F1-Score}}\\
\hline
 & 1DCNN & FRNN & LSTM & GRU & 1DCNN & FRNN & LSTM & GRU\\
\hline
{\bfseries 8h} &  &  &  &  &  &  &  &\\
CML & 0.68 $\pm$ 0.02 & 0.71 $\pm$ 0.02 & 0.71 $\pm$ 0.02 & 0.72 $\pm$ 0.02 & 0.86 $\pm$ 0.02 & 0.83 $\pm$ 0.02 & 0.83 $\pm$ 0.02 & 0.84 $\pm$ 0.02 \\
LML2 & 0.64 $\pm$ 0.04 & 0.69 $\pm$ 0.02 & 0.67 $\pm$ 0.04 & 0.67 $\pm$ 0.02 & 0.77 $\pm$ 0.04 & 0.81 $\pm$ 0.02 & 0.80 $\pm$ 0.03 & 0.80 $\pm$ 0.02 \\
LML4 & 0.63 $\pm$ 0.04 & 0.66 $\pm$ 0.05 & 0.68 $\pm$ 0.04 & 0.67 $\pm$ 0.03 & 0.76 $\pm$ 0.04 & 0.79 $\pm$ 0.03 & 0.80 $\pm$ 0.03 & 0.80 $\pm$ 0.03\\
LML8 & 0.58 $\pm$ 0.09 & 0.61 $\pm$ 0.07 & 0.61 $\pm$ 0.10 & 0.63 $\pm$ 0.06 & 0.71 $\pm$ 0.10 & 0.74 $\pm$ 0.07 & 0.75 $\pm$ 0.09 & 0.76 $\pm$ 0.06 \\
FL2 & 0.70 $\pm$ 0.01 & 0.69 $\pm$ 0.03 & 0.70 $\pm$ 0.02 & 0.70 $\pm$ 0.02 & 0.81 $\pm$ 0.01 & 0.81 $\pm$ 0.02 & 0.82 $\pm$ 0.02 & 0.82 $\pm$ 0.01\\
FL4 & 0.66 $\pm$ 0.03 & 0.67 $\pm$ 0.04 & 0.70 $\pm$ 0.03 & 0.64 $\pm$ 0.04 & 0.79 $\pm$ 0.03 & 0.80 $\pm$ 0.03 & 0.82 $\pm$ 0.02 & 0.78 $\pm$ 0.03\\
FL8 & 0.68 $\pm$ 0.01 & 0.67 $\pm$ 0.04 & 0.67 $\pm$ 0.05 & 0.69 $\pm$ 0.04 & 0.80 $\pm$ 0.01 & 0.80 $\pm$ 0.03 & 0.81 $\pm$ 0.03 & 0.81 $\pm$ 0.03\\
\hline
{\bfseries 16h} &  &  &  &  &  &  &  &\\
CML & 0.66 $\pm$ 0.04 & 0.72 $\pm$ 0.02 & 0.72 $\pm$ 0.02 & 0.71 $\pm$ 0.04 & 0.79 $\pm$ 0.03 & 0.83 $\pm$ 0.02 & 0.84 $\pm$ 0.02 & 0.83 $\pm$ 0.03\\
LML2 & 0.65 $\pm$ 0.04 & 0.68 $\pm$ 0.03 & 0.69 $\pm$ 0.03 & 0.69 $\pm$ 0.03 & 0.78 $\pm$ 0.03 & 0.80 $\pm$ 0.03 & 0.82 $\pm$ 0.02 & 0.81 $\pm$ 0.02\\
LML4 & 0.63 $\pm$ 0.05 & 0.65 $\pm$ 0.05 & 0.67 $\pm$ 0.04 & 0.65 $\pm$ 0.06 & 0.76 $\pm$ 0.04 & 0.78 $\pm$ 0.04 & 0.80 $\pm$ 0.03 & 0.78 $\pm$ 0.05\\
LML8 & 0.56 $\pm$ 0.07 & 0.59 $\pm$ 0.09 & 0.63 $\pm$ 0.09 & 0.64 $\pm$ 0.07 & 0.70 $\pm$ 0.07 & 0.73 $\pm$ 0.08 & 0.76 $\pm$ 0.08 & 0.77 $\pm$ 0.06\\
FL2 & 0.67 $\pm$ 0.02 & 0.67 $\pm$ 0.03 & 0.71 $\pm$ 0.02 & 0.69 $\pm$ 0.03 & 0.80 $\pm$ 0.02 & 0.79 $\pm$ 0.03 & 0.83 $\pm$ 0.02 & 0.81 $\pm$ 0.02\\
FL4 & 0.67 $\pm$ 0.04 & 0.70 $\pm$ 0.02 & 0.66 $\pm$ 0.07 & 0.68 $\pm$ 0.05 & 0.80 $\pm$ 0.03 & 0.82 $\pm$ 0.01 & 0.79 $\pm$ 0.05 & 0.81 $\pm$ 0.03\\
FL8 & 0.64 $\pm$ 0.04 & 0.70 $\pm$ 0.03 & 0.69 $\pm$ 0.02& 0.65 $\pm$ 0.05 & 0.78 $\pm$ 0.04 & 0.82 $\pm$ 0.02 & 0.82 $\pm$ 0.01 & 0.79 $\pm$ 0.04\\
\hline
{\bfseries 24h} &  &  &  &  &  &  &  &\\
CML & 0.67 $\pm$ 0.02 & 0.71 $\pm$ 0.03 & 0.72 $\pm$ 0.03 & 0.72 $\pm$ 0.02 & 0.79 $\pm$ 0.02 & 0.82 $\pm$ 0.03 & 0.83 $\pm$ 0.02 & 0.84 $\pm$ 0.02\\
LML2 & 0.68 $\pm$ 0.02 & 0.68 $\pm$ 0.04 & 0.68 $\pm$ 0.04 & 0.69 $\pm$ 0.03 & 0.80 $\pm$ 0.02 & 0.81 $\pm$ 0.03 & 0.80 $\pm$ 0.03 & 0.81 $\pm$ 0.02\\
LML4 & 0.63 $\pm$ 0.04 & 0.67 $\pm$ 0.03 & 0.66 $\pm$ 0.07 & 0.68 $\pm$ 0.03 & 0.77 $\pm$ 0.04 & 0.80 $\pm$ 0.03 & 0.79 $\pm$ 0.05 & 0.80 $\pm$ 0.02\\
LML8 & 0.60 $\pm$ 0.06 & 0.61 $\pm$ 0.10 & 0.63 $\pm$ 0.07 & 0.62 $\pm$ 0.09 & 0.74 $\pm$ 0.06 & 0.74 $\pm$ 0.09 & 0.76 $\pm$ 0.07 & 0.75 $\pm$ 0.08\\
FL2 & 0.66 $\pm$ 0.03 & 0.69 $\pm$ 0.02 & 0.71 $\pm$ 0.02 & 0.71 $\pm$ 0.02 & 0.78 $\pm$ 0.03 & 0.81 $\pm$ 0.02 & 0.83 $\pm$ 0.02 & 0.83 $\pm$ 0.01\\
FL4 & 0.67 $\pm$ 0.03 & 0.67 $\pm$ 0.06 & 0.70 $\pm$ 0.01 & 0.65 $\pm$ 0.08 & 0.80 $\pm$ 0.03 & 0.80 $\pm$ 0.04 & 0.83 $\pm$ 0.01 & 0.78 $\pm$ 0.06\\
FL8 & 0.67 $\pm$ 0.03 & 0.68 $\pm$ 0.04 & 0.66 $\pm$ 0.05  & 0.65 $\pm$ 0.05 & 0.80 $\pm$ 0.02 & 0.81 $\pm$ 0.03 & 0.80 $\pm$ 0.03 & 0.79 $\pm$ 0.04\\
\hline
{\bfseries 48h} &  &  &  &  &  &  &  &\\
CML & 0.68 $\pm$ 0.02 & 0.72 $\pm$ 0.03 & 0.71 $\pm$ 0.03 & 0.72 $\pm$ 0.03 & 0.81 $\pm$ 0.02 & 0.83 $\pm$ 0.02 & 0.82 $\pm$ 0.02 & 0.83 $\pm$ 0.02\\
LML2 & 0.64 $\pm$ 0.04 & 0.68 $\pm$ 0.04 & 0.70 $\pm$ 0.01 & 0.70 $\pm$ 0.02 & 0.77 $\pm$ 0.04 & 0.81 $\pm$ 0.03 & 0.82 $\pm$ 0.01 & 0.82 $\pm$ 0.01\\
LML4 & 0.61 $\pm$ 0.06 & 0.64 $\pm$ 0.04 & 0.68 $\pm$ 0.04 & 0.66 $\pm$ 0.04 & 0.74 $\pm$ 0.06 & 0.77 $\pm$ 0.04 & 0.80 $\pm$ 0.03 & 0.79 $\pm$ 0.04\\
LML8 & 0.58 $\pm$ 0.05 & 0.62 $\pm$ 0.05 & 0.63 $\pm$ 0.08 & 0.63 $\pm$ 0.08 & 0.72 $\pm$ 0.05 & 0.75 $\pm$ 0.05 & 0.75 $\pm$ 0.07 & 0.76 $\pm$ 0.07\\
FL2 & 0.68 $\pm$ 0.03 & 0.70 $\pm$ 0.03  & 0.68 $\pm$ 0.06 & 0.71 $\pm$ 0.03 & 0.80 $\pm$ 0.02 & 0.81 $\pm$ 0.02 & 0.80 $\pm$ 0.04 & 0.82 $\pm$ 0.02 \\
FL4 & 0.66 $\pm$ 0.03 & 0.69 $\pm$ 0.04 & 0.70 $\pm$ 0.03 & 0.67 $\pm$ 0.06 & 0.79 $\pm$ 0.03 & 0.81 $\pm$ 0.02 & 0.82 $\pm$ 0.02 & 0.80 $\pm$ 0.04\\
FL8 & 0.67 $\pm$ 0.03 & 0.68 $\pm$ 0.03 & 0.72 $\pm$ 0.02 & 0.67 $\pm$ 0.05 & 0.80 $\pm$ 0.01 & 0.81 $\pm$ 0.02 & 0.83 $\pm$ 0.01 & 0.81 $\pm$ 0.03\\
\hline
\end{tabular}
\vspace{-2mm}
\end{table*}

Recent research on ICU and in-hospital mortality prediction using MIMIC-III mainly focuses on AUC as an evaluation metric \cite{pattalung2021feature,lee2020federated}. Although AUC is widely used for evaluating classifiers built on imbalanced datasets, there is the drawback of the unreliability of the estimates when there is a low sample size of the minority class \cite{fernandez2018learning}. Thus, in this study, we focus our evaluation on AUPRC and F1-Score, which are common alternatives and better suited for highly imbalanced data.

We report the mean and standard deviation (std) of all approaches as follows: mean and std of all 5-fold models in CML, of k $\times$ 5-fold local models LML, and of all 5-fold global models in FL.

\smallskip
\noindent
\textbf{1DCNN vs FNN vs LSTM vs GRU}
Comparing the four classifiers, it is evident that all RNN classifiers, FRNN, LSTM, and GRU, are comparable. Nevertheless, on average, all three RNN classifiers are superior to 1DCNN, which underlines the fact that RNNs are designed for sequences, while CNNs are not capable of effectively learning temporal information.

\smallskip
\noindent
\textbf{Window Length}
All four classifiers (1DCNN, FRNN, LSTM, and GRU) have similar performance across all time windows (8h, 16h, 24h, and 48h) and approaches (CML, LML, and FL). This suggests that there is valuable information to be learned in all the windows and enough relevant information is also prevalent in the smaller time windows. This might be due to the fact that the most crucial information is observed shortly before the event of interest (ICU death/discharge). It could be argued that for dying patients, the shorter the window length, the higher the density of relevant information contained in vital signs and lab values. %For discharged patients the same reasoning cannot be applied. %However, an interesting observation is that a physicians' shift in the ICU usually lasts between 8 and 12 hours. Therefore, it can be speculated that it is indeed the data gathered during a shift that informs the discharge decision.

\smallskip
\noindent
\textbf{Number of Clients}
The experiments were performed with 2, 4, and 8 FL clients, simulating a set of independent (LML) or collaborating (FL) hospitals.
In LML, we notice that the performance continuously decreases with higher $K$ as the data available at each client decreases, and it could prove to be biased and/or too limited.
In FL, the results are comparable across all different number of clients, and there is no clear pattern to be observed. However, when the performance of FL with a certain number of clients is lower, the standard deviation is higher as well. This suggests that the data distribution influences the result in some rounds of the k-fold cross-validation, which could be solved by further optimization of the FL setup.

\smallskip
\noindent
\textbf{Comparison of CML, LML, and FL}
Our results illustrate that CML and FL have comparable performance for predicting ICU mortality. Both approaches perform well on the classification task considering the high class imbalance of only 9.75\% positive samples, with a baseline AUPRC of 0.10. Additionally, it can be verified that the behavior regarding the different window sizes matches between FL and CML. Extending the study by adding attention layers to the used classifiers could further verify whether the classifiers are learning the same patterns.
Furthermore, when comparing the predictive performance of LML and FL, it becomes apparent that FL performs considerably better than LML, which proves FL to be the better option over LML when data sharing amongst hospitals is not possible.

\smallskip
\noindent
\textbf{Limitations}
The data used (MIMIC-III) is from a single medical center, and selection bias is unavoidable. In addition, the data used is from the end of the ICU stay, knowing the time of death/discharge, and does not allow for early prediction. Thus, the retrospective nature of the analysis does not permit us to use this workflow within the scope of decision-support. Nevertheless, this study can be seen as the basis for further analysis of interpretability and feature importance.

\section{Conclusion}
\label{sec:summary_conclusions}

We presented a federated learning workflow for predicting ICU mortality using the MIMIC-III benchmark database. We compared the predictive performance of the proposed FL approach against LML and CML, using several sequential deep neural network classifiers (1DCNN, FRNN, LSTM, GRU), expanding windows of temporal data (8h, 16h, 24h, and 48h), and different numbers of FL clients (2, 4, and 8).
Our findings suggest that both CML and FL are comparable in terms of AUPRC and F1-Score. Additionally, FL is superior to LML, which is the only other alternative to guarantee data privacy. 

Since the main focus of this study was on comparing the different approaches, the FL setup has not been fully optimized. Thus, future work could involve experimentation on alternative design choices, such as using fixed mini-batch size, taking into consideration communication costs, and using real client/server FL architecture. Additional improvements could include exploring the general effect of local class distribution (fraction of deaths/dismissals per local client) within FL and employing rolling windows over each patient's history. % instead of a static window. 
Finally, integrating an interpretability method to determine the most important features in predicting ICU mortality in an FL approach is worth studying.

Overall, the FLICU workflow that we present in this study is for predicting ICU mortality using the MIMIC-III database. Nevertheless, our approach shows great promise to be easily extended not only to predict ICU mortality using different ICU databases but also on different clinical prediction tasks.

\bibliographystyle{IEEEtran}
\bibliography{references}

\end{document}